\documentclass[11pt]{article}

\usepackage[final]{acl}

\usepackage{times}
\usepackage{latexsym}

\usepackage[T1]{fontenc}

\usepackage[utf8]{inputenc}

\usepackage{microtype}

\usepackage{inconsolata}

\usepackage{graphicx}

\usepackage{hyperref}
\usepackage{amsmath,amssymb,amsthm}
\usepackage{booktabs}
\usepackage{multirow}
\usepackage{xcolor}
\usepackage{url}
\usepackage{algorithm}
\usepackage{algpseudocode}
\usepackage{bm}
\usepackage{pgfplots}
\usepackage{tabularx}
\usepackage{tikz}
\usetikzlibrary{shapes.geometric, arrows.meta, positioning, fit, backgrounds}
\pgfplotsset{compat=1.18}

\newtheorem{proposition}{Proposition}
\newtheorem{definition}{Definition}

\newcommand{\R}{\mathbb{R}}
\newcommand{\dtheta}{\Delta\bm{\theta}}
\newcommand{\dTheta}{\Delta\bm{\Theta}}
\newcommand{\base}{\mathrm{base}}
\newcommand{\eps}{\epsilon}
\definecolor{forgotcolor}{RGB}{220,80,60}
\definecolor{retaincolor}{RGB}{60,140,200}
\definecolor{servercolor}{RGB}{80,160,80}

\title{FIRM: Fisher-Informed Retention Merging 
       for Federated LLM Unlearning}
\author{Reeyansh Shah \quad Maram Hasan \quad Subhasis Chaudhuri \quad Biplab Banerjee \\
Indian Institute of Technology Bombay \\
\texttt{shah.reeyansh@gmail.com}}
\date{}

\begin{document}
\maketitle

\begin{abstract}
Federated LLM unlearning addresses targeted knowledge removal when
forget and retain data remain distributed across private clients and
are inaccessible to the server. The resulting locally trained
adapters encode conflicting objectives, making their aggregation
fundamentally different from conventional federated optimization. A
key failure mode is \emph{magnitude explosion}, whereby unnormalized
cross-client aggregation increases the merged update magnitude and
degrades retained utility under stronger unlearning. We propose \textbf{FIRM} (\textbf{Fi}sher-Informed \textbf{R}etention
\textbf{M}erging), a single-round framework for stable and
retention-aware server-side aggregation. FIRM uses client-local Fisher
statistics as a unified sensitivity signal to construct a
Fisher-normalized base merge and impose explicit control over
retain-sensitive parameters. Its second-order retention projection
admits a unique minimum-distance solution, while the corresponding
dual scales with the number of retain clients rather than the model
dimension. Experiments across WMDP, TOFU, and MUSE-HP using Zephyr-7B and
Llama-3-8B show utility gains at comparable forgetting on WMDP
and MUSE-HP, with metric-dependent trade-offs on TOFU.
Controlled comparisons support benefits of the combined base merge
and additional retention correction at selected operating points.
\end{abstract}

\section{Introduction}
\label{sec:intro}

The widespread deployment of large language models (LLMs) has made
selective knowledge removal practically important~\cite{maini2024tofu}. Deployed models may encode personal, copyrighted, or hazardous information that must be removed after training~\cite{li2024wmdp}. Regulatory
frameworks such as the GDPR~\cite{gdpr2016} and CCPA~\cite{ccpa2018}
further establish rights to erasure, while the cost of retraining
billion-parameter models makes targeted post-hoc removal the only
practical option in many settings. Despite this need, most LLM
unlearning methods assume centralized access to both the data to be
forgotten and representative retain data.

A critical but under-studied setting is \emph{federated LLM
unlearning}, where deletion-related data remain distributed across
private clients and cannot be transferred to the model provider under
localization constraints~\cite{liu2021federaser}. Unlike centralized
unlearning, the server cannot jointly access forget and retain data to
optimize or validate the resulting model. Unlike standard federated
learning, client objectives are not uniformly aligned toward minimizing
a shared task loss. Forget clients produce updates that suppress
targeted knowledge, whereas retain clients provide
preservation-oriented updates and parameter-sensitivity information.
The server must therefore arbitrate between destructive and protective
signals without access to either underlying
dataset~\cite{romandini2025survey}.

Recent work has addressed this problem through server-side aggregation
of locally trained adapters. FULM~\citep{zhong2025fulm}, for example,
applies TIES-based sign resolution~\citep{yadav2023ties} within client
groups, followed by unnormalized summation of the resulting
domain-level centroids. Although this preserves data locality and avoids
additional server-side training, the merged adapter can grow with
heterogeneous contributions rather than remain calibrated to their
collective importance. We refer to this failure mode as
\emph{magnitude explosion}. As the aggregate update grows, it may
suppress general capabilities together with targeted knowledge,
causing retained-utility collapse. Moreover, sign-based aggregation
captures directional agreement but not the \emph{degree} of parameter
importance to individual clients, and thus provides no explicit control
over retain-side damage.

Conventional adapter-merging methods do not directly resolve this
problem. TIES~\citep{yadav2023ties} and Fisher-weighted model
merging~\cite{matena2022merging} are primarily designed to combine
models optimized for compatible or cooperative objectives. Federated
unlearning instead combines updates with opposing functional roles: a
parameter may be sensitive on both forget and retain data, while
the corresponding task vectors favor opposing changes.
A suitable aggregation mechanism must therefore control both
the scale of the distributed update and its effect on
retain-sensitive parameters.

\noindent \textbf{Our Approach.} We introduce \textbf{FIRM}, a Fisher-informed framework for stable and
retention-aware federated LLM unlearning. FIRM uses empirical Fisher Information Matrix~(FIM) to quantify the parameter-wise importance of each client
update and employs it as a common signal throughout aggregation. This
allows the server to normalize heterogeneous unlearning contributions,
thereby preventing magnitude explosion, while simultaneously
protecting parameters that are important to retain clients. Hence, FIRM jointly controls update scale and a retain-side perturbation proxy
without requiring server-side access to either forget or retain data.

\paragraph{Contributions.}
\begin{enumerate}
\item We identify \textbf{magnitude explosion} as a key failure mode
in federated LLM unlearning, where unnormalized
cross-client aggregation amplifies the merged update and degrades
retained utility under stronger unlearning.

\item We propose \textbf{FIRM}, a single-round framework that uses
client-local Fisher information to construct a
\textbf{Fisher-normalized base merge}, enabling stable server-side
aggregation without access to forget or retain data.

\item We derive a \textbf{second-order retention-constrained
projection} that bounds a diagonal-Fisher perturbation proxy.
We establish the uniqueness of its minimum-distance solution
and obtain a dual formulation with one variable per retain
constraint, enabling tractable server-side optimization.\end{enumerate}

\section{Related Work}
\label{sec:related}

\noindent \textbf{Machine Unlearning for LLMs.}
Early methods apply gradient ascent directly to
increase the loss on forget examples~\cite{jang2022knowledge}, but can
be unstable and degrade general utility. More robust formulations
regularize the forget objective using retain data, including gradient
difference~\cite{liu2022continual} and KL-divergence penalties
~\cite{maini2024tofu}. Task Arithmetic~\cite{ilharco2023editing}
subtracts a fine-tuned task vector from the pretrained model, whereas
RMU~\cite{li2024wmdp} redirects selected intermediate representations
toward random targets. These methods assume centralized access to
forget and retain data and therefore do not directly address
distributed deletion requests under data-locality constraints.

\noindent \textbf{Adapter and Model Merging.}
The adoption of LoRA~\cite{hu2021lora} has motivated methods for
combining independently trained adapters.
TIES-Merging~\cite{yadav2023ties} mitigates interference through
trimming, sign-conflict resolution, and selective merging, while
KnOTS~\cite{stoica2024knot} aligns adapters in a shared subspace
using joint singular-value decomposition.
Fisher-weighted merging~\cite{matena2022merging} estimates
parameter importance using diagonal Fisher information and
combines model parameters through importance-weighted averaging,
giving greater influence to models with higher sensitivity
at each coordinate.
UniFLoW~\cite{liang2026uniflow} addresses aggregation inconsistency
in multimodal federated LoRA fine-tuning: averaging low-rank
factors separately generally differs from averaging their
induced weight updates. It averages one factor and analytically
reconstructs the other through regularized least squares.
CrossEarth-Gate~\cite{cao2026crossearthgate} uses empirical
Fisher information for remote-sensing adaptation, aggregating
parameter sensitivities into module-level scores to periodically
select which modules to train.
FIRM applies Fisher sensitivities to a different objective:
combining updates with opposing forget and retain roles.
It couples coordinate-wise Fisher-normalized merging with
a retention-constrained projection that explicitly bounds a Fisher-weighted perturbation proxy.

\noindent \textbf{Federated Unlearning.}
Earlier methods target conventional predictive models through
selective reconstruction~\cite{liu2021federaser}, rapid
retraining~\cite{liu2022right}, or class-discriminative activation
pruning~\cite{wang2022federated}. FedOSD~\cite{pan2025federated}
addresses forget--retain gradient conflict during iterative federated
unlearning by constructing an unlearning direction that avoids
interference with remaining-client gradients. It relies on repeated
client-gradient access, whereas FIRM operates in an LLM adapter
setting where each client transmits one task vector and Fisher
summary for training-free server-side aggregation.

Recent work extends federated unlearning to LLMs.
Oblivionis~\cite{zhang2026oblivionis} evaluates federated LLM
fine-tuning and targeted unlearning within an iterative framework,
evaluating combinations of federated optimizers and local unlearning
objectives. FULM~\cite{zhong2025fulm}, the closest
protocol-compatible framework, aggregates locally trained
unlearning and retention adapters through hierarchical TIES merging
followed by inter-group summation. However, its aggregation is
unnormalized and provides no explicit retain-side constraint. FIRM
addresses this aggregation gap through a single-round,
Fisher-normalized merge and a retention-constrained projection,
without server-side access to forget or retain data.
\begin{figure*}
    \centering
\includegraphics[width=1\linewidth]{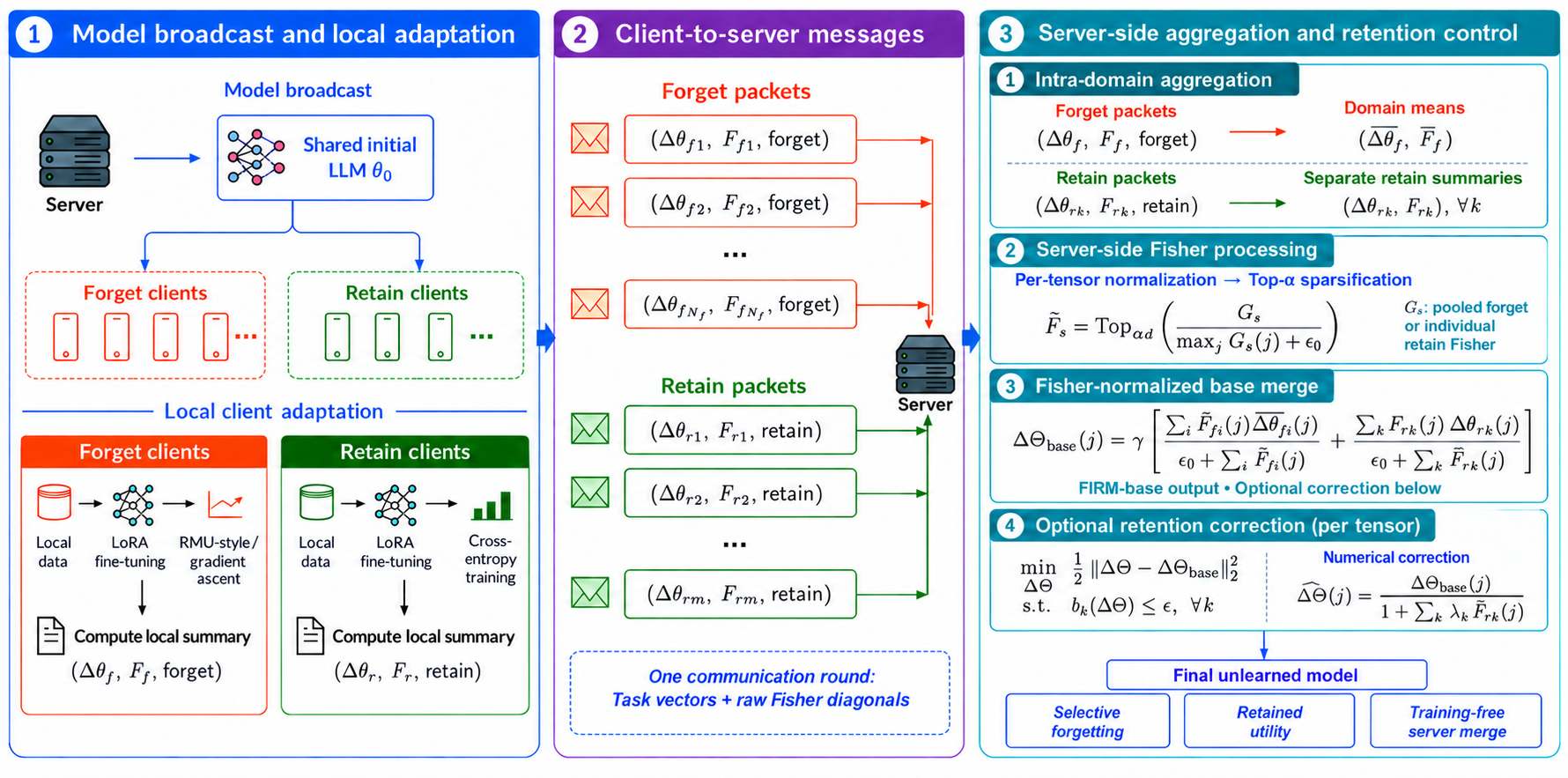}
\caption{Overview of FIRM. Clients transmit task vectors and raw
Fisher diagonals once. The server pools forget-domain summaries,
normalizes and sparsifies each Fisher tensor, and applies separately
normalized forget and retain merging with optional retention correction.}
\label{fig:overview}
\end{figure*}

\section{Problem Setup}
\label{sec:setup}

\noindent \textbf{Federated Unlearning Setting.}
We consider a central server hosting a shared initial LLM with
parameters $\bm{\theta}_0 \in \R^p$. A set of forget clients
$\mathcal{F}=\{f_i\}_{i=1}^{N_f}$ hold private datasets
$\mathcal{D}^{u}_{f_i}$ containing knowledge to be removed, while a
set of retain clients $\mathcal{R}=\{r_k\}_{k=1}^{m}$ hold private
datasets $\mathcal{D}^{r}_{r_k}$ representing knowledge to be
preserved. The server seeks an updated model
that achieves effective
forgetting while maintaining retained utility. No raw data is transmitted to the server at any point.


\vspace{0.15em}
\noindent \textbf{Local Adaptation via LoRA.}
Each client adapts a local model copy using LoRA~\cite{hu2021lora}.
For $W\in\R^{d_{\mathrm{out}}\times d_{\mathrm{in}}}$, the adapted
weight is $W+sBA$, where $s=\alpha_{\mathrm{LoRA}}/r$,
$B\in\R^{d_{\mathrm{out}}\times r}$,
$A\in\R^{r\times d_{\mathrm{in}}}$, and
$r\ll\min(d_{\mathrm{out}},d_{\mathrm{in}})$.
Task vectors concatenate $\mathrm{vec}(sBA)$ over adapted tensors.
Forget clients optimize an unlearning objective, whereas
retain clients optimize a preservation objective over their local
data.

\vspace{0.15em}

\noindent \textbf{Empirical Fisher Information.}
Each client $c$ estimates the diagonal empirical Fisher information
at shared pretrained reference parameters $\bm\theta_F$,
for $j=1,\ldots,p$:
\begin{equation}
F_c(j)=
\mathbb{E}_{x\sim\mathcal{D}_c}
\left[
\left(
\frac{\partial\mathcal{L}_{\mathrm{LM}}(x;\bm{\theta}_F)}
{\partial\theta_{F,j}}
\right)^2
\right].
\label{eq:fisher}
\end{equation}
Here $\mathcal L_{\mathrm{LM}}$ is causal-LM loss on the client's
data, including for RMU-trained forget clients. Fisher supplies
nonnegative sensitivities; signed task vectors supply direction.
The common reference $\bm\theta_F$ is the pretrained backbone;
for TOFU it precedes the expert adaptation used to initialize
client training (Appendix~\ref{app:impl}).


\section{Method}
\label{sec:method}

FIRM follows a single-round protocol: the server broadcasts the
pretrained model, clients perform local adaptation and transmit one
summary, and the server computes the final model without further
participation.  A retention-constrained projection then
limits updates along retain-sensitive parameters. Clients may assume forget or retain roles in different rounds; each client has a single role per round. Figure~\ref{fig:overview} illustrates the pipeline. Table~\ref{tab:notation} in Appendix~\ref{app:dual} summarizes the notation introduced in this section.

\subsection{Client-Side Sensitivity Encoding}
\label{ssec:client}

Each client locally adapts the pretrained model using LoRA and obtains
a task vector $\dtheta_c$ over the selected model layers. It transmits
$(\dtheta_c,F_c)$ once, using the raw empirical Fisher diagonal from
Eq.~\eqref{eq:fisher}. Only Fisher coordinates aligned with the task
vector enter server-side aggregation; raw client data remain local.

\subsection{Server-Side Aggregation Control}
\label{ssec:server}

\subsubsection{Intra-Domain Aggregation}
\label{ssec:intra}

When multiple forget clients share the same unlearning domain
$g$ (e.g., WMDP-Biology vs.\ WMDP-Cybersecurity), the server forms a domain-level centroid before cross-domain
aggregation:
\begin{equation}
  \overline{\dtheta}_f = \frac{1}{|\mathcal{C}_f|}
  \sum_{i \in \mathcal{C}_f} \dtheta_{f_i}, \qquad
  \bar F_f = \frac{1}{|\mathcal{C}_f|}
  \sum_{i \in \mathcal{C}_f} F_{f_i}.
\end{equation}

where $\mathcal{C}_f$ denotes forget clients from domain~$g$. Unlike
FULM's TIES sign-voting, FIRM uses mean aggregation and preserves
opposing signed contributions in the subsequent Fisher-normalized merge.
Retain summaries remain separate to preserve client-specific constraints.

\noindent \textbf{Server-Side Fisher Processing.}
After pooling, the server normalizes and sparsifies each Fisher
tensor independently. For a tensor with $d$ coordinates, let $G_s$
be a raw forget-domain centroid $\bar F_f$ or retain diagonal $F_{r_k}$:
\begin{equation}
  \widetilde F_s = \operatorname{Top}_{\alpha d}
  \left(\frac{G_s}{\max_j G_s(j)+\eps_0}\right).
  \label{eq:sparsification}
\end{equation}
Here $\alpha\in(0,1]$ sets the retained fraction; threshold ties
are kept (Appendix~\ref{app:impl}). We index processed forget-domain
sensitivities by $f_i$ below.




\subsubsection{Fisher-Weighted Base Merge}
\label{ssec:base}
Direct summation allows the aggregated update to grow with the number
of client contributions. FIRM instead constructs separate
Fisher-normalized forget and retain updates:
\begin{definition}[Fisher-Weighted Base Merge]
{\small
\begin{equation}
  \dTheta_{\mathrm{raw}}(j) =
  \frac{\sum_i \tilde{F}_{f_i}(j)\,\overline{\dtheta}_{f_i}(j)}
       {\sum_i \tilde{F}_{f_i}(j) + \eps_0}
  +
  \frac{\sum_k \tilde{F}_{r_k}(j)\,\dtheta_{r_k}(j)}
       {\sum_k \tilde{F}_{r_k}(j) + \eps_0},
  \label{eq:base}
\end{equation}
}
\end{definition}

where $\eps_0>0$ ensures numerical stability. As the number of clients
grows, the numerator and denominator scale together, preventing the
merged update from increasing proportionally with client count. This
\emph{self-normalizing} property mitigates magnitude explosion without
additional hyperparameters. 

\noindent \textbf{Global Scaling.} The base merge is \emph{scaled} as
\begin{equation}
    \dTheta_{\base}
    =
    \gamma\,\dTheta_{\mathrm{raw}},
    \label{eq:global_scale}
\end{equation}
where $\gamma>0$ controls the forget-retain trade-off.

\subsubsection{Retention-constrained Projection}
\label{ssec:qp}

Retention control is applied independently to each selected weight
tensor. Within this subsection, $\dTheta\in\R^d$ denotes its flattened
update; we suppress the tensor index. Each retain client has a budget
$\eps$ per tensor, not a global summed budget. We define the
\emph{second-order Fisher perturbation proxy}

\begin{equation}
b_k(\dTheta)
=
\frac{1}{2}
\sum_{j=1}^{d}
\widetilde F_{r_k}(j)\dTheta(j)^2,
\label{eq:damage}
\end{equation}

\noindent where $\tilde F_{r_k}(j)$ denotes the sparsified Fisher sensitivity of
parameter $j$ for retain client $r_k$. The proxy
$b_k(\dTheta)$ measures Fisher-weighted perturbation; squaring
$\dTheta(j)$ prevents directional cancellation and penalizes
retain-sensitive coordinates.

\begin{algorithm}[t]
\caption{FIRM: Fisher-Informed Retention Merging}
\label{alg:firm}
\begin{algorithmic}[1]
\Require Forget-client pairs
$\{(\dtheta_{f_i},F_{f_i})\}_{i=1}^{N_f}$;
retain-client pairs
$\{(\dtheta_{r_k},F_{r_k})\}_{k=1}^{m}$;
forget-client groups $\{\mathcal C_f\}$;
sparsity fraction $\alpha$;
global scaling $\gamma$;
retention tolerance $\eps$
\Statex \textbf{Client side}
\State \textbf{Summary Transmission:}\;
Send $(\dtheta_c,F_c)$ once to the server

\Statex \textbf{Server side}
\State \textbf{Intra-domain aggregation:}\;
Form forget-domain centroids
$\{(\overline{\dtheta}_{f_i},\bar F_{f_i})\}_{i=1}^{n}$
from the groups $\{\mathcal C_f\}$

\State \textbf{Fisher processing:}\;
Normalize and sparsify each pooled forget and separate retain
Fisher tensor via Eq.~\eqref{eq:sparsification}

\State \textbf{Base merge:}\;
Compute $\dTheta_{\mathrm{raw}}$ via Eq.~\eqref{eq:base}
\State \textbf{Scale:}\;
  $\dTheta_\base \leftarrow \gamma\,\dTheta_{\mathrm{raw}}$

\State \textbf{Retention correction (per tensor):}\;
Initialize $\bm\lambda=\mathbf 0$; increase multipliers for
violated constraints by bracketing and bisection until
feasibility or a stopping limit (Appendix~\ref{app:projection_dual}).
\Statex \hspace{1em}\textit{Recover:}\;
$\displaystyle
\widehat{\dTheta}(j) \leftarrow
\frac{\dTheta_{\base}(j)}
{1+\sum_k\lambda_k\tilde F_{r_k}(j)}$
\State \Return $\bm\theta_0$ plus the assembled tensor corrections
\end{algorithmic}
\end{algorithm}

\vspace{.3em}
\noindent \textbf{Second-Order Primal Program.}
For each tensor, the target projection is the closest update to the scaled base merge under the retain constraints:
\begin{equation}
\begin{aligned}
\min_{\dTheta\in\R^d}\quad
& \frac{1}{2}\|\dTheta-\dTheta_{\base}\|_2^2 \\
\mathrm{s.t.}\quad
& b_k(\dTheta)\leq\eps,
\qquad k=1,\ldots,m,
\end{aligned}
\label{eq:primal_second}
\end{equation}
where $\eps>0$ is the retention tolerance. Because each constraint $b_k(\dTheta)$ is a non-negatively weighted sum of squares, Eq.~\eqref{eq:primal_second} is strictly convex and satisfies Slater's condition at $\dTheta = \mathbf{0}$, ensuring strong duality and a unique global optimum.

\vspace{0.3em}
\noindent \textbf{Coordinate-Wise Shrinkage and Dual Reduction.} Introducing the dual Lagrange multipliers
$\bm{\lambda}=[\lambda_1,\ldots,\lambda_m]^\top\geq\mathbf{0}$,
the stationarity condition yields, for $j=1,\ldots,d$,
\begin{equation}
  \dTheta^*(\bm{\lambda})(j)
  =
  \frac{\dTheta_{\base}(j)}
  {1+\sum_{k}\lambda_k\tilde F_{r_k}(j)}.
  \label{eq:primal_shrinkage}
\end{equation}
Thus, parameters with larger retain sensitivity undergo stronger
shrinkage, while insensitive parameters remain closer to the base
merge.

The dual has $m$ variables per tensor; evaluating its objective
and gradient costs $O(md)$. Experiments use scalar bisection for
$m=1$ and sequential multiplier increases for $m>1$.
The latter checks feasibility, not full optimality; solver details
and the exact dual derivation are in
Appendix~\ref{app:projection_dual}. All computation remains
server-side, with no additional communication rounds.



\subsection{Theoretical Guarantees}
\label{ssec:theory}

The following guarantees characterize the exact optimum of the per-tensor program, rather than every numerical solver output:

\begin{proposition}[Proxy Retention Bound]
The optimal solution $\dTheta^*$ of the second-order primal program~\eqref{eq:primal_second} satisfies $b_k(\dTheta^*) \le \eps$ for all retain clients $k = 1, \ldots, m$.
\end{proposition}

\begin{proposition}[Minimal Modification]
$\dTheta^*$ is the unique $\ell_2$-closest tensor update to $\dTheta_{\base}$ satisfying all $m$ retention-proxy constraints.
\end{proposition}

\begin{proposition}[Scalable Dual Solution]
The dual of problem~\eqref{eq:primal_second} has $m$ variables
per tensor, independent of its dimension $d$. Strong duality holds
for $\eps>0$; projected gradient ascent with an appropriate step
size recovers the dual optimum. Evaluation cost still depends on $d$.
\end{proposition}

\noindent Algorithm~\ref{alg:firm} summarises the FIRM pipeline.







  

\section{Experiments}
\label{sec:experiments}

\subsection{Setup}
\label{ssec:setup}

\paragraph{Models.}
Primary experiments use \textbf{Zephyr-7B-$\beta$}~\cite{tunstall2023zephyr},
a 7B-parameter instruction-tuned model based on Mistral-7B, loaded
in 4-bit NF4 quantisation~\cite{dettmers2024qlora}.
Robustness experiments use
\textbf{Llama-3-8B-Instruct}~\cite{touvron2023llama} on the
MUSE-HP benchmark. We use LoRA rank $r = 16$ with scaling coefficient $\alpha_{\mathrm{LoRA}} = 32$
throughout.
Fisher diagonals are computed from a separately loaded bfloat16 copy
to estimate sensitivities without 4-bit quantisation
(Appendix~\ref{app:impl}).

\vspace{0.3em}
\noindent \textbf{WMDP Setup (Zephyr-7B).}
We use six forget clients: three hold disjoint shards of
WMDP-Biology (1,273 questions, $\approx$424 per client), and
three hold disjoint shards of WMDP-Cybersecurity (1,987 questions,
$\approx$662 per client). A single retain client trains on a
random 600-sample subset of MMLU~\cite{hendrycks2021mmlu}. Forget
adapters use an RMU-style representation-misdirection
objective~\cite{li2024wmdp}, updating down-projection weights
in layers 5--7. Forget clients optimize the forgetting term;
retention is handled by the separate retain client and server merge.

\vspace{0.3em}
\noindent \textbf{TOFU Setup (Zephyr-7B).}
The model is first fine-tuned on the full TOFU dataset (200 fictional
authors, 4,000 QA pairs) to establish a knowledgeable baseline.
Five forget clients each hold a disjoint shard of the forget01 split
(40 QA pairs, 2 authors total).
A single retain client trains on 200 samples from the retain90 split.
Forget adapters use gradient ascent on all linear layers.

\vspace{0.3em}
\noindent \textbf{MUSE-HP Setup (Llama-3-8B).}
Three forget clients each train on a disjoint shard of 1,676
passages ($\approx$4,096 chars each) obtained from the MUSE-Books
raw splits~\cite{nguyen2024muse}.
Forget evaluation: 100 KnowMem QA pairs from the forget\_qa split,
measured by ROUGE-L, while
Utility is evaluated on MMLU (14,042 questions).
Forget adapters use gradient ascent on all linear layers.

\vspace{0.3em}
\noindent \textbf{Baselines.}
\textbf{AVG}: uniform mean of all task vectors.
\textbf{SUM}: direct summation of forget task vectors.
\textbf{TIES}~\cite{yadav2023ties}: sign-vote merging of forget updates.
\textbf{FULM}~\cite{zhong2025fulm}: hierarchical TIES intra-cluster,
summation inter-cluster.
All baselines use identical training conditions and hardware.
Numerical discrepancies from originally reported FULM results
may arise from undisclosed implementation details; we report our
controlled reimplementation for fair comparison.

\vspace{0.3em}
\noindent \textbf{Metrics.}
\textbf{Bio$\downarrow$} and \textbf{Cyber$\downarrow$} denote
WMDP-Biology and WMDP-Cybersecurity accuracy, respectively (lower = better;
random chance 25\%).
\textbf{HP$\downarrow$}: MUSE-HP ROUGE-L, and
\textbf{Forget$\downarrow$}: TOFU forget-set ROUGE-L.
\textbf{Retain$\uparrow$}, \textbf{Real Authors$\uparrow$}, and
\textbf{Real World$\uparrow$} are the TOFU retention metrics, while
\textbf{Utility$\uparrow$} denotes MMLU accuracy. 


\vspace{0.3em}
\noindent \textbf{Hyperparameters.}
FIRM and FULM use independent $\gamma$ sweeps; other baseline
settings are detailed in Appendix~\ref{app:impl}.
We set $\eps_0=10^{-8}$ and use Top-10\% Fisher
sparsification ($\alpha=0.1$), selected through the ablation.
We distinguish \textbf{FIRM-base}, which uses Fisher-normalized
merging without retention projection ($\eps=\infty$), from
\textbf{FIRM-QP}, which additionally applies the projection
with finite $\eps$. Tables~\ref{tab:sweep}--\ref{tab:llama}
evaluate FIRM-base; the
retention-control experiments separately evaluate FIRM-QP.
The matched WMDP QP comparisons use $\eps=4.1\times10^{-5}$
and a tighter budget of $10^{-5}$; multi-client budgets are
specified in Table~\ref{tab:multi_retain}. We allow up to
500 solver iterations. All experiments ran on NVIDIA
RTX PRO 6000 Blackwell Max-Q GPUs (96\,GB VRAM).

\subsection{Main Results: WMDP (Zephyr-7B)}
\label{ssec:main}

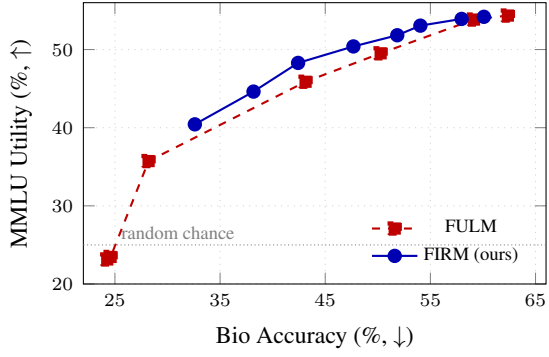
\begin{figure}[t]
\centering
\begin{tikzpicture}
\begin{axis}[
    width=\columnwidth,
    height=5.3cm,
    xlabel={Bio Accuracy (\%, $\downarrow$)},
    ylabel={MMLU Utility (\%, $\uparrow$)},
    xmin=22, xmax=66,
    ymin=20, ymax=56,
    xtick={25,35,45,55,65},
    ytick={20,30,40,50},
    grid=major,
    grid style={dotted, gray!35},
    tick label style={font=\scriptsize},
    label style={font=\small},
    legend style={font=\scriptsize, draw=none, fill=none},
    legend pos=south east,
]
\addplot[color=red!75!black, mark=square*, mark size=2.3pt,
         dashed, thick] coordinates {
    (62.37,54.35)(59.15,53.84)(50.27,49.52)(43.13,45.88)
    (28.20,35.73)(24.59,23.46)(24.27,23.14)
};
\addlegendentry{FULM}
\addplot[color=blue!70!black, mark=*, mark size=2.3pt, thick]
  coordinates {
    (60.09,54.19)(57.97,53.92)(54.05,53.06)(51.85,51.83)
    (47.68,50.40)(42.42,48.29)(38.18,44.62)(32.60,40.43)
};
\addlegendentry{FIRM (ours)}
\addplot[gray, densely dotted, thin, domain=22:66] {25};
\node[font=\scriptsize, gray] at (axis cs:31,26.5) {random chance};
\end{axis}
\end{tikzpicture}
\caption{Forgetting--utility trade-off on WMDP with Zephyr-7B.
Each point denotes a scaling value $\gamma$. The discussed
comparable-forgetting pair favors FIRM; connecting segments
visualize the sweep and do not establish uniform dominance.}
\label{fig:pareto}
\end{figure}

\paragraph{Forgetting--Utility Pareto Frontier.}
Table~\ref{tab:sweep} reports WMDP-Bio and WMDP-Cyber accuracies
(lower indicates stronger forgetting) and MMLU Utility
(higher indicates better retained performance) across the tested
$\gamma$ values. FIRM's Utility declines gradually from 54.19\% to
40.43\% as $\gamma$ increases from 0.5 to 1.2, whereas FULM shows a
sharp degradation for $\gamma \geq 0.2$, with Utility falling to
23--36\%. At comparable Bio forgetting depth
(Bio $\approx 43\%$), FIRM retains 48.29\% Utility versus 45.88\%
for FULM, while achieving similar Cyber accuracy
(31.91\% versus 31.76\%). This gives a \textbf{2.41\,pp Utility
improvement} at comparable forgetting depth. The scale-tuned
controls in Section~\ref{ssec:ablations} further assess the combined
merge beyond global rescaling. Figure~\ref{fig:pareto} plots the sweep.

\begin{table}[t]
\centering
\scriptsize
\setlength{\tabcolsep}{10.5pt}
\caption{WMDP sweep on Zephyr-7B.
FIRM uses Fisher-weighted merging without QP
($\eps=\infty$); matched-forgetting QP comparisons
are reported in Table~\ref{tab:qp_matched}. $\downarrow/\uparrow$: lower/higher
is better. $\dagger$: Utility collapse ($\leq25\%$, near random).
\textbf{Bold}: FIRM utility values for ease of comparison.}
\label{tab:sweep}
\begin{tabular}{llrrr}
\toprule
\textbf{Method} & $\bm{\gamma}$ &
\textbf{Bio$\downarrow$} & \textbf{Cyber$\downarrow$} &
\textbf{Util$\uparrow$} \\
\midrule
Pretrained & --- & 63.08 & 40.11 & 54.46 \\
\midrule
FULM & 0.05 & 62.37 & 40.01 & 54.35 \\
FULM & 0.10 & 59.15 & 38.95 & 53.84 \\
FULM & 0.15 & 50.27 & 36.03 & 49.52 \\
FULM & 0.17 & 43.13 & 31.76 & 45.88 \\
FULM & 0.20 & 28.20 & 26.17 & 35.73 \\
FULM & 0.30 & 24.59 & 24.66 & 23.46$^\dagger$ \\
FULM & 0.40 & 24.27 & 26.02 & 23.14$^\dagger$ \\
\midrule
FIRM (ours) & 0.50 & 60.09 & 38.90 & \textbf{54.19} \\
FIRM (ours) & 0.60 & 57.97 & 38.65 & \textbf{53.92} \\
FIRM (ours) & 0.70 & 54.05 & 38.05 & \textbf{53.06} \\
FIRM (ours) & 0.80 & 51.85 & 37.29 & \textbf{51.83} \\
FIRM (ours) & 0.90 & 47.68 & 35.03 & \textbf{50.40} \\
FIRM (ours) & 1.00 & 42.42 & 31.91 & \textbf{48.29} \\
FIRM (ours) & 1.10 & 38.18 & 30.80 & \textbf{44.62} \\
FIRM (ours) & 1.20 & 32.60 & 28.84 & \textbf{40.43} \\
\bottomrule
\end{tabular}
\end{table}

\begin{table}[t]
\centering
\scriptsize
\setlength{\tabcolsep}{6pt}
\caption{WMDP results at the listed evaluated settings
  (Zephyr-7B). Scale-controlled averaging is compared separately
  in Table~\ref{tab:average_firm}. Bold marks the highest utility
  among the listed settings.}
\label{tab:comparison}
\begin{tabular}{lcccc}
\toprule
\textbf{Method} & \textbf{Setting} &
\textbf{Bio$\downarrow$} & \textbf{Cyber$\downarrow$} &
\textbf{Util$\uparrow$} \\
\midrule
AVG         & scale=1.00    & 24.59 & 25.87 & 24.18 \\
SUM         & scale=0.05    & 38.41 & 30.60 & 43.06 \\
TIES        & topk=0.50     & 24.74 & 25.11 & 22.80 \\
FULM        & $\gamma=0.17$ & 43.13 & 31.76 & 45.88 \\
FIRM (ours) & $\gamma=1.00$ & 42.42 & 31.91 &
              \textbf{48.29} \\
\bottomrule
\end{tabular}
\end{table}

\vspace{0.3em}
\paragraph{Comparison with Merging Baselines.}
Table~\ref{tab:comparison} reports the listed evaluated settings.
AVG averages six forget and one retain vector at unit scale;
SUM and TIES use only forget vectors. FIRM achieves the highest
utility among these settings (48.29\%). These rows differ in
inputs and tuning; Table~\ref{tab:average_firm} separately tests
FIRM against independently scale-tuned averaging.

\subsection{Near-iid Unlearning: TOFU (Zephyr-7B)}
\label{ssec:tofu}
Table~\ref{tab:tofu} evaluates FIRM on TOFU~\cite{maini2024tofu},
a near-iid fictional-author unlearning benchmark with closely
related forget and retain distributions. This makes targeted
forgetting challenging without degrading retained
fictional-author knowledge. FULM achieves stronger forgetting
at aggressive scales, but with substantial retention loss:
from $\gamma=1.0$ to $\gamma=2.5$, retain ROUGE-L drops from
95.51\% to 67.84\%, while real-author performance declines
from 44.65\% to 23.20\%. FIRM improves forgetting more gradually,
maintaining retain ROUGE-L above 94\% across the evaluated
scales. At comparable forgetting depth, FIRM at $\gamma=16.0$
achieves lower forget ROUGE-L than FULM at $\gamma=1.0$
(70.00\% vs.\ 71.85\%), with comparable retain performance
(95.55\% vs.\ 95.51\%) and a 2.50-point improvement in
real-author ROUGE-L (47.15\% vs.\ 44.65\%). This provides
a favorable operating point for targeted forgetting and
preservation of author knowledge. The trade-off remains
metric-dependent: FULM achieves higher MMLU accuracy
(37.35\% vs.\ 36.44\%) and real-world ROUGE-L
(75.93\% vs.\ 72.76\%) at these settings.
\begin{table}[t]
\centering
\scriptsize
\setlength{\tabcolsep}{6pt}
\caption{TOFU unlearning on Zephyr-7B (base merge without QP).
  Fgt$\downarrow$: forget ROUGE-L (lower = better).
  Ret$\uparrow$, RA$\uparrow$, RW$\uparrow$: retain, real authors,
  real world ROUGE-L.
  Util$\uparrow$: MMLU accuracy. Bold marks the highest retention
  and utility scores among the unlearned models.}
\label{tab:tofu}
\begin{tabular}{llrrrrr}
\toprule
\textbf{Method} & $\bm{\gamma}$ &
\textbf{Fgt$\downarrow$} & \textbf{Ret$\uparrow$} &
\textbf{RA$\uparrow$} & \textbf{RW$\uparrow$} &
\textbf{Util$\uparrow$} \\
\midrule
Expert baseline & --- & 96.74 & 94.64 & 47.15 & 77.78 & 39.49 \\
\midrule
FULM & 1.0 & 71.85 & 95.51 & 44.65 & 75.93 & 37.35 \\
FULM & 1.5 & 61.68 & 91.51 & 42.45 & 70.80 & 35.94 \\
FULM & 2.0 & 51.84 & 78.16 & 34.03 & 66.97 & 32.60 \\
FULM & 2.5 & 43.71 & 67.84 & 23.20 & 64.34 & 33.78 \\
\midrule
FIRM & 1.0 & 90.15 & \textbf{96.75} & \textbf{49.15} & \textbf{77.64} & \textbf{38.40} \\
FIRM & 2.0 & 81.27 & 94.26 & 45.40 & 74.62 & 37.09 \\
FIRM & 4.0 & 78.89 & 95.08 & 45.90 & 74.90 & 36.48 \\
FIRM & 6.0 & 76.47 & 95.69 & 46.65 & 74.76 & 36.23 \\
FIRM & 8.0 & 73.87 & 95.50 & 47.15 & 72.62 & 36.70 \\
FIRM & 12.0 & 71.16 & 95.82 & 46.65 & 71.91 & 36.68 \\
FIRM & 16.0 & 70.00 & 95.55 & 47.15 & 72.76 & 36.44 \\
\bottomrule
\end{tabular}
\end{table}


\begin{table}[t]
\centering
\small
\caption{MUSE-HP unlearning on Llama-3-8B (base merge without QP).
  HP$\downarrow$: ROUGE-L on forget QA (lower = better forgetting).
  Util$\uparrow$: MMLU accuracy.
  $\dagger$: utility collapse ($\leq25\%$, near random).
  Bold marks the highest utility at zero HP ROUGE-L.}
\label{tab:llama}
\begin{tabular}{llrr}
\toprule
\textbf{Method} & $\bm{\gamma}$ &
\textbf{HP$\downarrow$} & \textbf{Util$\uparrow$} \\
\midrule
Pretrained & --- & 50.57 & 61.12 \\
\midrule
FULM & 0.10 &  5.52 & 59.59 \\
FULM & 0.20 &  0.00 & 57.52 \\
FULM & 0.30 &  0.00 & 53.95 \\
FULM & 0.50 &  0.00 & 35.67 \\
FULM & 0.70 &  0.00 & 22.95$^\dagger$ \\
\midrule
FIRM (ours) & 0.30 & 45.38 & 60.87 \\
FIRM (ours) & 0.50 & 28.41 & 60.38 \\
FIRM (ours) & 0.70 & 11.71 & 60.30 \\
FIRM (ours) & 1.00 &  4.57 & 59.78 \\
FIRM (ours) & 1.20 &  2.37 & 59.43 \\
FIRM (ours) & 1.50 &  0.33 & 58.72 \\
FIRM (ours) & 1.60 &  0.00 & \textbf{58.57} \\
\bottomrule
\end{tabular}
\end{table}

\subsection{Robustness: MUSE-HP (Llama-3-8B)}
\label{ssec:llama}

Table~\ref{tab:llama} evaluates FIRM on a second model family and a
different unlearning domain, demonstrating generalization across
architectures and knowledge types. At zero forget-QA ROUGE-L
(HP $=0\%$), FULM achieves its best Utility of 57.52\% at
$\gamma=0.2$, whereas FIRM reaches \textbf{58.57\%} at
$\gamma=1.6$, yielding a 1.05\,pp gain among the evaluated
zero-ROUGE-L settings. This compares utility at equal measured
forgetting; numerical $\gamma$ intervals are method-dependent
and do not by themselves establish reduced sensitivity to scaling.
Cross-model evaluations on MUSE-HP/Zephyr-7B and WMDP/Llama-3-8B
provide further comparisons in Appendix~\ref{app:cross_model}.

\subsection{Ablation Studies}
\label{ssec:ablations}

We examine three design choices at fixed $\gamma = 1.0$ on WMDP.
Results are reported in Table~\ref{tab:ablation}.

\begin{table}[t]
\centering
\small
\setlength{\extrarowheight}{0pt}
\caption{Ablation study on WMDP (Zephyr-7B, $\gamma=1.0$).
  $\dagger$: catastrophic utility collapse ($\leq$25\%).}
\label{tab:ablation}
\begin{tabular}{lrrr}
\toprule
\textbf{Configuration} &
\textbf{Bio$\downarrow$} & \textbf{Cyber$\downarrow$} &
\textbf{Util$\uparrow$} \\
\midrule
FIRM-base ($\alpha=0.10$)     & 42.42 & 31.91 & 48.29 \\
A1: No sum-norm                & 63.00 & 40.01 & 54.37 \\
A2: Uniform weights            & 24.82 & 25.87 & 24.54$^\dagger$ \\
A3: Top-20\% ($\alpha=0.20$)   & 34.64 & 29.59 & 42.92 \\
A3: Full Fisher ($\alpha=1.0$) & 24.82 & 25.87 & 24.54$^\dagger$ \\
\midrule
Pretrained                     & 63.08 & 40.11 & 54.46 \\
\bottomrule
\end{tabular}
\end{table}

\vspace{0.25em}
\noindent \textbf{A1 --- Unnormalized Fisher Aggregation.}
Using an unnormalized raw-Fisher weighted sum produces
near-identity behavior at the tested scale: Bio accuracy is
63.00\% and MMLU utility is 54.37\%, close to the pretrained
model. This illustrates the importance of calibrating the
weighted update, but does not establish that an independently
rescaled variant would behave identically.

\vspace{0.25em}
\noindent \textbf{A2 --- Uniform Weighting.}
Replacing Fisher weights with uniform weights over all
coordinates yields Bio accuracy of 24.82\% and MMLU utility
of 24.54\% at $\gamma=1.0$. This demonstrates substantial
utility degradation under the tested configuration.
However, the replacement also changes sparse support and
update magnitude, so this comparison does not isolate
continuous Fisher weighting.

\vspace{0.25em}
\noindent \textbf{A3 --- Fisher Sparsification.}
Increasing the retained Fisher fraction from Top-10\% to
Top-20\% reduces MMLU utility from 48.29\% to 42.92\%;
using the full diagonal reduces it to 24.54\%. These results
favor Top-10\% at the tested scale. Because sparsification
changes both coordinate support and the resulting update,
they do not establish low-sensitivity noise as the cause
of degradation.

\vspace{0.25em}
\noindent \textbf{Discussion.}
These fixed-scale ablations support the chosen combination
of normalization, Fisher weighting, and sparsification.
Their effects on support and update magnitude are coupled,
motivating the scale-tuned comparisons below and the
support/norm-matched control in Appendix~\ref{app:weight_control}.

\vspace{0.25em}
\noindent \textbf{FIRM versus Scale-Tuned Averaging.}
For identical inputs, $\mathrm{AVG}=\mathrm{SUM}/N$, so
$\gamma$ absorbs this global rescaling. Table~\ref{tab:average_firm}
compares FIRM-base with two averaging baselines on shared
regenerated WMDP checkpoints, using separate $\gamma$ values
and no QP. Dense uniform averaging uses the six-forget mean
plus the retain update; FIRM gains 2.756 MMLU points at
0.236/0.101-point Bio/Cyber gaps. FULM-MEAN preserves
clustering and intra-cluster TIES, replacing inter-cluster
SUM with MEAN. Against this variant, FIRM gains 0.612 MMLU
points at identical Cyber accuracy and a 0.157-point Bio gap.
These comparisons support the combined FIRM design beyond
global rescaling at the reported operating points.
Protocols, additional comparisons, and the full-sweep summary are in
Appendix~\ref{app:controlled_comparisons}.

\begin{table}[t]
\centering
\small
\setlength{\tabcolsep}{3pt}
\caption{FIRM-base versus scale-tuned averaging on shared
WMDP checkpoints, without QP. Each block compares closely
matched forgetting; all accuracies are in \%.}
\label{tab:average_firm}
\begin{tabular}{lrrrr}
\toprule
Method & $\gamma$ & Bio$\downarrow$ & Cyber$\downarrow$ & MMLU$\uparrow$ \\
\midrule
Uniform avg. & 0.30 & 45.80 & 34.78 & 48.58 \\
FIRM-base & 1.00 & 45.56 & 34.88 & \textbf{51.33} \\
\midrule
FULM-MEAN & 0.35 & 58.21 & 38.70 & 53.77 \\
FIRM-base & 0.60 & 58.05 & 38.70 & \textbf{54.38} \\
\bottomrule
\end{tabular}
\end{table}

\subsection{Retention Control and Dual Scalability}

\label{ssec:qp_verify}

\noindent \textbf{Retention Control at Comparable Forgetting.}
A fixed-$\gamma$ comparison cannot establish whether the
retention projection improves utility beyond simply reducing
the update scale. We therefore compare FIRM-QP with
FIRM-base across independently evaluated $\gamma$ values
on shared regenerated WMDP checkpoints. QP/base matches
minimize the maximum absolute Bio/Cyber accuracy gap,
without using MMLU for selection.
Table~\ref{tab:qp_matched} reports representative comparisons.

At $\gamma=1.5$, FIRM-QP improves MMLU by 0.669 percentage
points over FIRM-base at $\gamma=1.1$, with identical Bio
accuracy and a 0.201-point Cyber gap. At $\gamma=2.2$
versus $1.5$, FIRM-QP achieves slightly stronger forgetting
in both domains and improves MMLU by 1.723 points. These
results demonstrate a benefit from retention-aware
projection beyond scalar shrinkage at these operating points.

\begin{table}[t]
\centering
\small
\setlength{\tabcolsep}{3pt}
\caption{Representative retention-control comparisons on
WMDP (Zephyr-7B), using shared regenerated checkpoints.
F: FIRM-base ($\eps=\infty$);
Q: FIRM-QP with $\eps=4.1\times10^{-5}$;
T: FIRM-QP with $\eps=10^{-5}$.
Parentheses give the first/second method's $\gamma$.
Accuracy gaps are absolute; $\Delta$MMLU is first minus
second. All differences are in percentage points.}
\label{tab:qp_matched}
\begin{tabular}{lrrr}
\toprule
\textbf{Pair ($\gamma$)} & \textbf{Bio gap}
& \textbf{Cyber gap} & \textbf{$\Delta$MMLU} \\
\midrule
Q/F (1.5/1.1) & 0.000 & 0.201 & +0.669 \\
Q/F (2.2/1.5) & 0.393 & 0.302 & +1.723 \\
\midrule
T/Q (1.1/1.0) & 0.079 & 0.101 & +0.249 \\
T/Q (1.5/1.4) & 0.471 & 0.252 & +0.349 \\
T/Q (2.4/2.3) & 0.393 & 0.302 & +1.453 \\
\bottomrule
\end{tabular}
\end{table}

Tightening the tolerance to $\eps=10^{-5}$ provides
additional control over the trade-off. For the three
reported comparisons, the nearest Bio and Cyber matches
select the same standard-QP reference, and tighter
projection improves MMLU by 0.249--1.453 points.
These are local improvements rather than uniform
dominance across the sweep. Additional comparisons and
the fixed-$\gamma$ tolerance analysis appear in
Appendix~\ref{app:controlled_comparisons}.

\begin{table}[t]
\centering
\scriptsize
\setlength{\tabcolsep}{1.8pt}
\renewcommand{\arraystretch}{1.08}
\caption{Retention-constraint checks with heterogeneous retain
clients on WMDP. Damage is measured per tensor; Bio and Utility
are reported as no-QP/QP. For $m=1,2,3$, respectively,
$\eps\approx(9.98,5.68,6.30)\times10^{-5}$, calibrated
separately as 0.3 times the mean no-QP damage across
retain clients and tensors.}
\label{tab:multi_retain}
\begin{tabular}{c|c|cc|c|c|c}
\toprule
$m$
& Active
& \multicolumn{2}{c|}{$\max_k b_k$}
& Time
& Bio$\downarrow$
& Util$\uparrow$ \\
& $\lambda_k$
& Before
& After
& (s)
& no-QP/QP
& no-QP/QP \\
\midrule
1 & 1/1
& $5.7{\times}10^{-4}$
& $1.0{\times}10^{-4}$
& 0.77
& 42.97/49.73
& 48.16/51.69 \\

2 & 2/2
& $3.6{\times}10^{-4}$
& $5.7{\times}10^{-5}$
& 0.99
& 42.81/49.80
& 47.63/51.66 \\

3 & 3/3
& $4.2{\times}10^{-4}$
& $6.3{\times}10^{-5}$
& 1.66
& 42.50/50.98
& 47.67/52.12 \\
\bottomrule
\end{tabular}
\end{table}

\noindent \textbf{Multi-Client Constraint Checks.}
Table~\ref{tab:multi_retain} reports positive multipliers for all
retain clients and reduced per-tensor Fisher damage, with higher
utility and modest server-side cost. These measurements assess
constraint satisfaction and overhead; they do not certify
minimum-distance optimality for $m>1$.

\section{Analysis}
\label{sec:analysis}

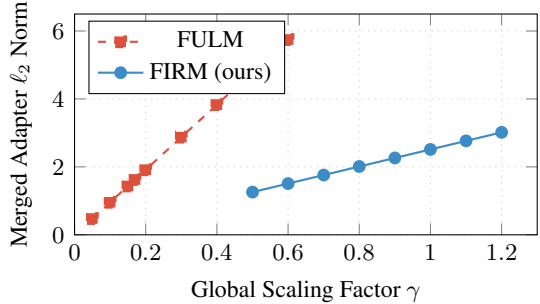
\begin{figure}[t]
\centering
\begin{tikzpicture}
\begin{axis}[
  width=\columnwidth, height=4.5cm,
  xlabel={Global Scaling Factor $\gamma$},
  ylabel={Merged Adapter $\ell_2$ Norm},
  xmin=0.0, xmax=1.3, ymin=0, ymax=6.5,
  xtick={0.0,0.2,0.4,0.6,0.8,1.0,1.2},
  legend pos=north west, legend style={font=\small},
  grid=major, grid style={dotted, gray!40},
  tick label style={font=\small}, label style={font=\small},
]
\addplot[color=forgotcolor, mark=square*, mark size=2pt,
         thick, dashed] coordinates {
  (0.05,0.479)(0.10,0.958)(0.15,1.437)(0.17,1.629)
  (0.20,1.916)(0.30,2.875)(0.40,3.833)(0.50,4.791)(0.60,5.749)
};
\addlegendentry{FULM}
\addplot[color=retaincolor, mark=*, mark size=2pt, thick]
  coordinates {
  (0.50,1.257)(0.60,1.508)(0.70,1.759)(0.80,2.011)
  (0.90,2.262)(1.00,2.513)(1.10,2.765)(1.20,3.016)
};
\addlegendentry{FIRM (ours)}
\end{axis}
\end{tikzpicture}
\caption{Merged adapter $\ell_2$ norm versus $\gamma$ over three WMDP
LoRA tensors. FIRM has smaller norms at equal $\gamma$.
This illustrates method-dependent update scaling; performance
comparisons require tuning $\gamma$ and matching forgetting.}
\label{fig:magnitude}
\end{figure}

\paragraph{How Normalization Controls Update Magnitude.}
Consider $n$ heterogeneous forget-domain centroids.
FULM's unnormalized inter-domain summation can accumulate aligned
contributions, increasing the update norm as $n$ grows.
In FIRM's base merge~\eqref{eq:base}, adding the $k$-th centroid
increases both the numerator and denominator by
$\tilde{F}_{f_k}(j)\overline{\dtheta}_{f_k}(j)$ and $\tilde{F}_{f_k}(j)$,
respectively.
Each role-specific term is a coordinate-wise weighted average with
weight sum at most one. Its absolute value is therefore bounded by
the largest contributing update magnitude at that coordinate,
avoiding growth caused solely by summing more terms.
Figure~\ref{fig:magnitude} shows the norm difference across scaling
factors. Table~\ref{tab:domain_scaling} confirms this trend as the
number of heterogeneous centroids increases: FULM's norm rises more
rapidly and its utility collapses, whereas FIRM's norm begins to
stabilize and retains substantially higher utility.

\begin{table}[t]
\centering
\small
\setlength{\tabcolsep}{3pt}
\caption{Effect of increasing the number of heterogeneous forget-domain
centroids $n$ on WMDP: Bio, then Cyber, then Chemistry, with
one forget client per domain and one retain client.
Scales remain fixed at $\gamma=0.15$ (FULM) and $1.0$ (FIRM),
with no active QP. Norms cover layers 5--7.}
\label{tab:domain_scaling}
\begin{tabular}{c|cc|cc}
\toprule
$n$
& \multicolumn{2}{c|}{Adapter norm}
& \multicolumn{2}{c}{Utility$\uparrow$} \\
& FULM & FIRM & FULM & FIRM \\
\midrule
1 & 0.73 ($\times1.0$) & 1.65 ($\times1.0$) & 54.22 & 52.68 \\
2 & 1.44 ($\times2.0$) & 2.59 ($\times1.6$) & 43.31 & 47.44 \\
3 & 1.86 ($\times2.6$) & 2.78 ($\times1.7$) & 24.03 & 39.10 \\
\bottomrule
\end{tabular}
\end{table}

\paragraph{Computational Cost.}
FIRM adds a one-time client-side Fisher estimation step.
In the measured WMDP configuration, Fisher estimation
takes 15.96 minutes per client. Server-side merging takes
217 seconds, compared with 547 seconds for FULM, while
the retention correction adds 1.2 seconds.
Appendix~\ref{app:efficiency} reports the runtime, memory,
and communication breakdown.


\section{Conclusion}
\label{sec:conclusion}
We introduced \textbf{FIRM}, a single-round framework for federated
LLM unlearning that combines local forget and retain updates without
server access to client data. FIRM couples Fisher-normalized merging
with retention control based on a diagonal-Fisher perturbation proxy.
The per-tensor projection formulation has a unique minimum-distance
optimum and an $m$-variable dual. Experiments across two model families
show utility gains at comparable forgetting on WMDP and MUSE-HP,
with metric-dependent trade-offs on TOFU. Scale-tuned comparisons
support benefits of the combined Fisher-based design beyond global
rescaling, while matched-forgetting comparisons demonstrate additional
benefits from retention correction at selected operating points.
Future work will examine larger models, mixed-role clients, broader
heterogeneity, and the relationship between Fisher-proxy control and
actual retention.

\section*{Limitations}

FIRM constrains a second-order empirical Fisher proxy rather than the
exact retain-loss change. Although this provides non-negative,
parameter-wise retention control, its fidelity depends on how well the
diagonal empirical Fisher represents the local retain-loss curvature.
Establishing tighter bounds between the proxy and the true loss change
remains an important theoretical direction.

Our evaluation covers two model families across WMDP, TOFU, and
MUSE-HP, including cross-model tests on WMDP and MUSE-HP. Validation
beyond the 8B scale, on additional model families, and under broader
client and domain heterogeneity remains future work.

The current protocol assigns each client a single role for a given
request, either contributing a forget update or a retain summary.
Extending FIRM to mixed-role clients that simultaneously hold both
deletion-requested and retained data is left for future work.

Our protocol assumes a trusted central
server and honest-but-curious clients. Robustness to manipulated Fisher statistics, poisoned task vectors,
client dropout, and fully decentralized aggregation remains open.
\bibliography{references}

\appendix
\section{Notations}
\label{app:dual}

Table~\ref{tab:notation} summarizes the notation used in our paper.

\begin{table}[h]
\centering
\scriptsize
\setlength{\tabcolsep}{3pt}
\renewcommand{\arraystretch}{0.9}
\begin{tabular}{lp{0.68\columnwidth}}
\toprule
\textbf{Symbol} & \textbf{Meaning} \\
\midrule
$\Delta\theta_c$ & Concatenated $\mathrm{vec}(sBA)$, with $s=\alpha_{\mathrm{LoRA}}/r$ \\
$\bm\theta_0,\bm\theta_F$ & Initial model and pretrained Fisher reference \\
$\alpha,\alpha_{\mathrm{LoRA}}$ & Fisher retained fraction and LoRA scaling coefficient \\
$F_c(j)$ & Raw empirical Fisher diagonal for parameter $j$ of client $c$ \\
$\tilde F_s(j)$ & Per-tensor max-normalized, Top-$\alpha d$ Fisher sensitivity \\
$\mathcal{C}_f$ & Forget clients sharing unlearning domain $g$ \\
$\overline{\Delta\theta}_f,\ \overline{F}_f$ & Per-domain centroids (means over $\mathcal{C}_f$) \\
$N_f$ & Number of forget clients before domain pooling \\
$n$ & Number of forget-domain centroids in the merge \\
$m$ & Number of retain clients (constraints per tensor) \\
$d$ & Number of coordinates in a selected weight tensor \\
$\Delta\Theta_{\mathrm{raw}}$ & Unscaled Fisher-weighted merge (Eq.~\eqref{eq:base}) \\
$\Delta\Theta_{\mathrm{base}}$ & Scaled merge $\gamma\Delta\Theta_{\mathrm{raw}}$ (Eq.~\eqref{eq:global_scale}) \\
$\gamma$ & Global update scaling factor \\
$\epsilon_0$ & Numerical stabilizer in the base-merge denominator ($10^{-8}$) \\
$\epsilon$ & Retention-proxy budget per client per tensor \\
$b_k(\Delta\Theta)$ & Fisher perturbation proxy for client $k$ in one tensor \\
$S(\lambda)$ & Diagonal shrinkage matrix, $I+\sum_k\lambda_kD_k$ \\
$\lambda_k$ & Dual multiplier for retain client $k$'s constraint \\
$\Delta\Theta^*,\widehat{\Delta\Theta}$ & Exact per-tensor optimum and numerical correction \\
\bottomrule
\end{tabular}

\caption{Main notation used in FIRM.}
\label{tab:notation}
\end{table}

\subsection{Cross-Model Robustness}
\label{app:cross_model}

To test robustness beyond the primary model--benchmark pairings, we
evaluate MUSE-HP on Zephyr-7B and WMDP on Llama-3-8B. Because
$\gamma$ is method-specific, comparisons are made at similar
forgetting levels rather than identical values. The comparisons
below show utility gains at selected operating points, with
the remaining differences in forgetting reported explicitly.

\paragraph{MUSE-HP / Zephyr-7B.}
FIRM at $\gamma=0.30$ reaches HP ROUGE-L of 0.06\% with
50.98\% MMLU, compared with 0.81\% and 47.35\% for FULM
at $\gamma=0.07$: lower measured recall and 3.63 points higher
utility. At exactly zero HP ROUGE-L, FIRM at $\gamma=0.40$
retains 49.34\% MMLU versus 35.37\% for FULM at $\gamma=0.10$.
Zero ROUGE-L concerns the evaluated answers, not proof of
complete knowledge removal.

\paragraph{WMDP on Llama-3-8B.}
Table~\ref{tab:appendix_llama_wmdp} exhibits the same pattern. At a
comparable forgetting level, FULM obtains Bio/Cyber accuracies of
30.87\%/27.73\% with 29.65\% MMLU, whereas FIRM obtains
33.70\%/27.08\% while retaining 36.58\% MMLU. Thus, FIRM preserves
6.93\,pp more utility, with weaker Bio forgetting by 2.83 points
and stronger Cyber forgetting by 0.65 points. This is a trade-off
with nonzero matching gaps, not an equal-forgetting comparison.

These sweeps use an inactive QP ($\epsilon=999$), so they
evaluate the combined base merge without projection. They do
not isolate continuous Fisher weighting. Numerical $\gamma$
interval widths are not comparable measures of robustness;
the evidence comes from the measured forgetting--utility pairs.
\begin{table}[!h]
\centering
\small
\setlength{\tabcolsep}{4pt}
\caption{Supplementary sweep results on Zephyr-7B / MUSE-HP.}
\label{tab:appendix_zephyr_muse}
\begin{tabular}{lccr}
\toprule
\textbf{Method} & $\bm{\gamma}$ & \textbf{HP} $\downarrow$ & \textbf{MMLU} $\uparrow$ \\
\midrule
Pretrain & -- & 44.07 & 54.46 \\
FULM & 0.005 & 43.43 & 54.29 \\
FULM & 0.010 & 41.85 & 54.19 \\
FULM & 0.020 & 41.01 & 53.88 \\
FULM & 0.030 & 30.25 & 53.51 \\
FULM & 0.050 & 10.23 & 52.21 \\
FULM & 0.070 & 0.81 & 47.35 \\
FULM & 0.100 & 0.00 & 35.37 \\
FIRM & 0.050 & 41.43 & 54.32 \\
FIRM & 0.100 & 34.37 & 53.82 \\
FIRM & 0.200 & 12.23 & 53.45 \\
FIRM & 0.300 & 0.06 & 50.98 \\
FIRM & 0.400 & 0.00 & 49.34 \\
FIRM & 0.500 & 0.00 & 48.17 \\
\bottomrule
\end{tabular}
\end{table}

\begin{table}[!h]
\centering
\small
\setlength{\tabcolsep}{4pt}
\caption{Supplementary sweep results on Llama-3-8B / WMDP.}
\label{tab:appendix_llama_wmdp}
\begin{tabular}{lccrr}
\toprule
\textbf{Method} & $\bm{\gamma}$ & \textbf{Bio} $\downarrow$ & \textbf{Cyber} $\downarrow$ & \textbf{MMLU} $\uparrow$ \\
\midrule
Pretrain & -- & 70.93 & 43.53 & 61.12 \\
FULM & 0.05 & 71.41 & 43.63 & 61.10 \\
FULM & 0.10 & 70.78 & 42.68 & 60.82 \\
FULM & 0.15 & 70.31 & 42.07 & 60.30 \\
FULM & 0.20 & 69.21 & 41.67 & 59.36 \\
FULM & 0.30 & 61.82 & 38.45 & 56.62 \\
FULM & 0.35 & 54.75 & 33.32 & 53.19 \\
FULM & 0.40 & 39.04 & 28.89 & 44.50 \\
FULM & 0.45 & 30.87 & 27.73 & 29.65 \\
FULM & 0.50 & 25.22 & 27.38 & 25.04 \\
FIRM & 0.70 & 70.54 & 43.18 & 60.63 \\
FIRM & 0.90 & 69.99 & 42.33 & 59.93 \\
FIRM & 1.00 & 69.99 & 41.87 & 59.88 \\
FIRM & 1.10 & 69.99 & 41.92 & 59.33 \\
FIRM & 1.20 & 69.21 & 41.57 & 58.89 \\
FIRM & 1.50 & 65.20 & 39.26 & 57.31 \\
FIRM & 1.75 & 60.96 & 36.34 & 53.58 \\
FIRM & 2.00 & 46.35 & 31.56 & 48.45 \\
FIRM & 2.25 & 33.70 & 27.08 & 36.58 \\
FIRM & 2.50 & 26.00 & 26.98 & 28.49 \\
\bottomrule
\end{tabular}
\end{table}
\section{Dual Derivation and Numerical Retention Correction}
\label{app:projection_dual}

We derive the exact per-tensor projection and distinguish it from
the numerical correction used in Algorithm~\ref{alg:firm}.
For tensor $\ell$, let $d_\ell$ be its number of coordinates.
The full objective separates across tensors when each has its
own constraints $b_{k,\ell}\leq\eps$. The summed proxy for a
client is then bounded by $T\eps$ for $T$ constrained tensors,
not by $\eps$. Below, we suppress $\ell$ and write $d=d_\ell$.

\paragraph{Setup.}
For retain client $k$, let $D_k = \mathrm{diag}(\tilde F_{r_k}) \succeq 0$
be the diagonal matrix formed from its sparsified, normalised Fisher
diagonal. The retain damage proxy is
\begin{equation}
  b_k(\dTheta) = \tfrac12\,\dTheta^\top D_k \dTheta .
\end{equation}
The target problem is
\begin{equation*}
\begin{aligned}
\min_{\dTheta\in\R^d}\quad
& \tfrac12\left\|\dTheta-\dTheta_{\base}\right\|_2^2 \\
\text{s.t.}\quad
& b_k(\dTheta) \leq \eps,
\qquad k=1,\ldots,m .
\end{aligned}
\tag{P}
\label{eq:primal}
\end{equation*}
The objective is strictly convex and each constraint is a convex
quadratic, so \eqref{eq:primal} is a convex program with a unique
solution. Since $\dTheta=\mathbf 0$ gives $b_k(\mathbf 0)=0<\eps$
for every $k$, Slater's condition holds, strong duality applies,
and the KKT conditions fully characterise the optimum.

\paragraph{Lagrangian.}
With multipliers $\lambda_k \geq 0$,
{\small
\begin{align}
  \mathcal L
  &= \tfrac12\|\dTheta-\dTheta_\base\|_2^2
     + \sum_{k=1}^m \lambda_k\bigl(b_k(\dTheta)-\eps\bigr)\\
  &= \tfrac12\,\dTheta^\top S(\bm\lambda)\,\dTheta
     - \dTheta^\top\dTheta_\base + c(\bm\lambda),
\end{align}
}
where $S(\bm\lambda) = I + \sum_k \lambda_k D_k$ is diagonal with
entries $S(\bm\lambda)_{jj} = 1+\sum_k\lambda_k\tilde F_{r_k}(j)$,
and $c(\bm\lambda)$ collects the terms not depending on $\dTheta$.
Because $\lambda_k\ge 0$, we always have $S(\bm\lambda)\succeq I$.

\paragraph{Solving for $\dTheta$.}
Setting $\nabla_\dTheta \mathcal L = 0$ gives
$S(\bm\lambda)\dTheta=\dTheta_\base$, hence
{\small
\begin{equation}
  \dTheta^*(\bm\lambda)(j)
  = \frac{\dTheta_\base(j)}
         {1+\sum_{k} \lambda_k \tilde F_{r_k}(j)} .
  \label{eq:shrinkage}
\end{equation}
}

For fixed $\bm\lambda$, this is the Lagrangian minimizer;
it is the constrained optimum only at optimal multipliers.
Algorithm~\ref{alg:firm} uses this shrinkage form with numerical
stabilization and returns $\widehat{\dTheta}$. Nonnegative
multipliers ensure coordinate magnitudes never increase;
this controls the Fisher proxy, not the exact retain loss.

\paragraph{Dual function.}
Plugging \eqref{eq:shrinkage} back in gives the dual objective
$g(\bm\lambda) = \mathcal L(\dTheta^*(\bm\lambda),\bm\lambda)$.
Because $\mathcal L$ is linear in $\bm\lambda$ for any fixed
$\dTheta$, and $\dTheta^*(\bm\lambda)$ minimises $\mathcal L$ for
each $\bm\lambda$, its derivative is simply
\begin{equation}
  \frac{\partial g}{\partial\lambda_k}(\bm\lambda)
  = b_k\bigl(\dTheta^*(\bm\lambda)\bigr) - \eps .
  \label{eq:dual_grad}
\end{equation}
As a pointwise infimum of affine functions of $\bm\lambda$,
$g$ is concave. Convergence of projected gradient ascent also
requires an appropriate step size, as specified below.

\paragraph{Dual problem and update rule.}
The dual is
\begin{equation}
  \max_{\bm\lambda\ge \mathbf 0} \; g(\bm\lambda),
\end{equation}
and can be solved by projected gradient ascent using \eqref{eq:dual_grad}:
\begin{equation}
  \bm\lambda \leftarrow \max\Bigl(\mathbf 0,\;
  \bm\lambda - \eta\bigl(\eps\mathbf 1
  - b(\dTheta^*(\bm\lambda))\bigr)\Bigr),
  \label{eq:dual_update}
\end{equation}
This theoretical update allows multipliers to both increase and
decrease. It differs from the increase-only bisection procedure
used for the reported multi-client runs.

\paragraph{Optimality conditions.}
The KKT conditions for \eqref{eq:primal} are
\begin{equation}
\dTheta^*-\dTheta_\base
+\Bigl(\textstyle\sum_k \lambda_k^*D_k\Bigr)\dTheta^*=0,
\end{equation}
\vspace{-1.5em}
\begin{align}
b_k(\dTheta^*) &\leq \eps, \qquad \forall k, \\
\lambda_k^* &\geq 0, \qquad \forall k, \\
\lambda_k^*\bigl(b_k(\dTheta^*)-\eps\bigr)
&=0, \qquad \forall k.
\end{align}

Since \eqref{eq:primal} is convex and Slater's condition holds,
these are necessary and sufficient, so any $(\dTheta^*,\bm\lambda^*)$
satisfying them is the global solution.

\paragraph{Single retain client ($m=1$).}
Here $b_1(\dTheta^*(\lambda_1))$ decreases smoothly and
monotonically as $\lambda_1$ grows from $0$ to $\infty$, going
from $b_1(\dTheta_\base)$ down to $0$. So:
\begin{itemize}
  \item if $b_1(\dTheta_\base)\le\eps$, the base merge is already
  safe: $\lambda_1^*=0$ and $\dTheta^*=\dTheta_\base$;
  \item otherwise there is a unique $\lambda_1^*>0$ solving
  $b_1(\dTheta^*(\lambda_1))=\eps$, found quickly by bisection.
\end{itemize}
Scalar bisection therefore recovers the single-client projection
to numerical precision. The multi-client implementation is
described separately below.

\paragraph{Scalability and an exact dual solver.}
Each tensor has $m$ dual variables. Evaluating the objective
and gradient requires $O(md)$ operations; storing its Fisher
diagonals requires $O(md)$ memory. Across tensors, evaluation
cost scales as $O(m\sum_\ell d_\ell)$, so a small dual dimension
does not make computation independent of model size.
With $f_j=[\tilde F_{r_1}(j),\ldots,\tilde F_{r_m}(j)]^\top$, the
negative dual Hessian is
\begin{equation}
 -\nabla^2 g(\bm\lambda)
 =\sum_{j=1}^{d}
 \frac{\dTheta_\base(j)^2 f_j f_j^\top}
 {(1+\bm\lambda^\top f_j)^3}.
\end{equation}
Consequently, a gradient Lipschitz bound is
\begin{equation}
 L=\sum_{j=1}^{d}\dTheta_\base(j)^2\|f_j\|_2^2.
 \label{eq:lipschitz}
\end{equation}
For $L>0$, projected gradient ascent with $0<\eta\leq1/L$
converges to a dual optimum, yielding the unique primal optimum.
If $L=0$, the base update has zero proxy damage and is feasible.

\paragraph{Numerical procedure used in the experiments.}
The implementation initializes separate multipliers to zero for
each selected tensor. For each violated constraint, it brackets
a larger multiplier by doubling and performs 60 bisection steps,
holding the other multipliers fixed. It stops when all computed
damage values are at most $\eps+\tau$, where
$\tau=\max(10^{-4}\eps,10^{-12})$, or when the multiplier-change
or iteration limit is reached. Experiments allow up to 500 outer
iterations; stopping at a limit alone does not certify feasibility.
The recovered update uses an additional denominator stabilizer
$\eps_0$ and is cast to the task-vector dtype. Constraint checks
on the delivered update should account for this rounding.

For $m=1$, bisection solves the scalar boundary equation.
For $m>1$, the implemented routine only increases multipliers.
Later increases can make an earlier constraint slack while its
multiplier remains positive, violating complementary slackness.
Thus, satisfying the proxy budgets alone does not certify the
minimum-distance solution. The multi-client experiment reports
empirical constraint satisfaction and cost, not a KKT certificate.
An exact-solver implementation would need to permit multiplier
decreases and verify all KKT residuals.

\paragraph{Guarantees and their scope.}
The exact per-tensor optimum satisfies every proxy budget and is
the unique closest update to the base merge. Its dual has $m$
variables. These mathematical properties do not establish
multi-client optimality for the increase-only numerical routine,
nor do the proxy budgets bound the exact retain-loss change.

\begin{table}[t]
\centering
\scriptsize
\setlength{\tabcolsep}{2.4pt}
\renewcommand{\arraystretch}{1.05}
\begin{tabularx}{\columnwidth}{p{0.57\columnwidth}X}
\toprule
\textbf{Component (per client unless stated)}
& \textbf{Time / Size} \\
\midrule
Adapter training
& 0.63 min \\
bf16 model load for Fisher
& 0.11 min \\
Fisher estimation, 20 batches
& 15.96 min$^\ast$ \\
Peak GPU memory: training / Fisher
& 19.7 / 38.4$^\ast$ GB \\
Task-vector checkpoint
& 336 MiB \\
Dense Fisher on adapter support (derived)
& 336 MiB values \\
Full Fisher checkpoint in profiling run
& 13,562 MiB \\
Server: FIRM merge / QP solve
& 217 s / 1.2 s \\
Server: FULM merge
& 547 s \\
\bottomrule
\end{tabularx}
\caption{Client-side and server-side efficiency breakdown for FIRM.
$^\ast$The measurement keeps the 4-bit training model resident while
loading the bf16 Fisher copy. These measurements include both
resident model copies; an isolated Fisher pass was not measured.}
\label{tab:gdcb_fisher_cost}
\end{table}

\section{Implementation Details}
\label{app:impl}

\paragraph{Fisher Computation.}
Fisher diagonals are computed at pretrained reference weights
$\bm\theta_F$ on a fresh bfloat16 model copy, avoiding 4-bit
NF4 quantisation noise. For WMDP and MUSE-HP, this is also
the client initialization $\bm\theta_0$. For TOFU, client training
starts from the expert-adapted model, while Fisher estimation
loads the original backbone without that expert adapter.
The Fisher batch size is one sequence, with 20 batches per client
and gradients reset before each backward pass. Each sequence uses
the causal-LM loss averaged over non-padding target tokens.
Its gradient is squared element-wise before averaging across
the 20 batches: $\widehat F_c(j)=\frac{1}{20}\sum_{b=1}^{20}
[\partial\mathcal L_{\mathrm{LM}}(x_b;\bm\theta_F)/\partial\theta_{F,j}]^2$.
Thus, token-loss averaging precedes squaring, while cross-example
averaging follows squaring. Squared BF16 gradients are accumulated
in FP32 and the average is stored in BF16. Sequences are truncated
to 512 tokens; batches follow the supplied data order, cycling
if fewer than 20 examples are available. Only entries aligned
with adapter task-vector keys are used during merging.

\paragraph{Server-Side Fisher Processing.}
The server first averages raw Fisher tensors within each forget
domain. It then divides each pooled forget tensor and each separate
retain tensor by its own maximum plus $\eps_0$, and applies the
Top-$\alpha d$ threshold in Eq.~\eqref{eq:sparsification}.
Top-$\alpha d$ keeps values at least as large as the
$\max(1,\lfloor\alpha d\rfloor)$-th largest entry, including ties.
Pooling precedes both normalization and sparsification; these
operations do not generally commute. Multi-retain experiments
disable retain-client grouping to preserve separate constraints.

\paragraph{Adapter Configuration.}
All adapters use LoRA $r=16$, $\alpha_{\mathrm{LoRA}}=32$. WMDP adapters target
\texttt{down\_proj} in layers 5--7; TOFU and MUSE-HP adapters target
all seven projection modules across all layers.

\paragraph{Original Baseline Settings.}
The original SUM sweep used scales
$\{0.05,0.1,0.2,0.4,0.6,1.0\}$; TIES used support fractions
$\{0.1,0.2,0.3,0.5\}$ without an additional scale sweep.
Table~\ref{tab:comparison} reports SUM at 0.05, TIES at 0.5,
and one unit-scaled AVG run. These are evaluated settings,
not the outcome of a common scale-matched selection rule.

\subsection{Efficiency Analysis}
\label{app:efficiency}

We instrumented one complete WMDP client run and the server-side
pipeline. Table~\ref{tab:gdcb_fisher_cost} reports the corresponding
runtime, memory, and communication costs.

\paragraph{Computational Cost.}
FIRM introduces a one-time client-side Fisher estimation step after
local adapter training. With 20 Fisher batches, this is the dominant
client-side cost, but it is performed independently and can run in
parallel across clients. Server-side computation remains comparatively
light: the FIRM merge takes 217\,s, compared with 547\,s for FULM,
while the constrained dual projection adds only 1.2\,s. Thus, FIRM's
additional optimization cost is concentrated in local Fisher
estimation rather than the server-side projection.

\paragraph{Memory Cost.}
Local adapter training requires 19.7\,GB of peak GPU memory, whereas
the Fisher pass reaches 38.4\,GB because the measurement retains the
4-bit training model while loading a separate bf16 copy for
gradient-stable Fisher estimation. An isolated Fisher pass was not
measured. The server operates only on transmitted
task vectors and Fisher summaries and therefore requires no access to
client data or full client models.

\paragraph{Communication Cost.}
Each client sends a task vector and raw Fisher summary once;
server-side projection needs no further communication. The profiled
task-vector artifact occupies 336\,MiB. The original full Fisher
checkpoint occupies 13,562\,MiB; restricting its bf16 values to
task-vector support gives a derived dense payload of 336\,MiB.
These are artifact sizes and payload estimates, not measured network
traffic. Top-10\% sparsification occurs after server-side pooling
and therefore does not reduce the evaluated upload payload.

\paragraph{Compute Budget.}
NVIDIA RTX PRO 6000 Blackwell Max-Q (96\,GB VRAM).

\section{Additional Controlled Comparisons}
\label{app:controlled_comparisons}

\subsection{Scale-Tuned Uniform Baseline}
The dense Fisher-free baseline uses
\begin{equation}
\Delta_{\mathrm{uniform}} =
\gamma\left(\frac{1}{6}\sum_{i=1}^{6}
\Delta_i^{\mathrm{forget}}+\Delta^{\mathrm{retain}}\right).
\end{equation}
It applies neither masking nor projection. Both methods use
shared regenerated checkpoints and full Bio/Cyber/MMLU
evaluation sets (1,273/1,987/14,042 questions). This checkpoint
set differs from the original main-result and ablation runs.
The sweep comprises 17 FIRM-base and 33 uniform settings,
including uniform refinement over $[0.20,0.40]$ in steps of 0.01.
Table~\ref{tab:uniform_selected} gives representative pairs
across forgetting levels, including an equal-accuracy pair.
Residual gaps are explicit; the stronger-forgetting example
exceeds 0.5 points in Bio. The comparison supports the combined
design, not continuous Fisher weighting alone or uniform
superiority across operating points. No multi-seed uncertainty
is estimated.

This uniform baseline differs from the FULM-MEAN variant
evaluated separately below, which retains TIES aggregation.
Using all 17 FIRM and 33 uniform settings, we also construct
observed-step envelopes: the highest observed MMLU under each
Bio or Cyber accuracy ceiling, without interpolation or
extrapolation. Over common accuracy ranges of
25.22--61.59\% (Bio) and 25.97--39.41\% (Cyber),
range-normalized areas favor FIRM by 1.268 and 2.326
percentage points, respectively. These grid-dependent summaries
do not establish statistical significance or uniform dominance.

\begin{table}[!htbp]
\centering
\small
\setlength{\tabcolsep}{3pt}
\caption{Representative scale-tuned comparisons on WMDP. F: FIRM-base; U: dense uniform. Parentheses give first/second $\gamma$. Accuracy gaps are absolute; $\Delta$MMLU is F minus U. Differences are percentage points.}
\label{tab:uniform_selected}
\begin{tabular}{lrrr}
\toprule
Pair ($\gamma$) & Bio gap & Cyber gap & $\Delta$MMLU \\
\midrule
F/U (0.4/0.10) & 0.000 & 0.000 & -0.007 \\
F/U (0.6/0.20) & 0.236 & 0.050 & +0.413 \\
F/U (1.0/0.30) & 0.236 & 0.101 & +2.756 \\
F/U (1.3/0.35) & 0.550 & 0.151 & +2.471 \\
\bottomrule
\end{tabular}
\end{table}

\subsection{FULM with Inter-Cluster Averaging}
\label{app:fulm_mean}
Using the same regenerated checkpoints and evaluation sets,
we preserve FULM's similarity-based clustering (threshold
0.6) and intra-cluster TIES (Top-30\%), replacing only
inter-cluster SUM with MEAN. The three clusters contain
the Bio clients, Cyber clients, and retain client, respectively;
all three centroids enter the mean. We evaluate seven scales,
$\gamma\in\{0.30,0.35,0.40,0.45,0.50,0.55,0.60\}$.
Each is matched to the 17 FIRM-base settings by minimizing
the maximum absolute Bio/Cyber gap, breaking ties by the
sum of gaps and then lower reference $\gamma$, without
using MMLU. The 0.35/0.60 pair in
Table~\ref{tab:average_firm} is the only base comparison
with both gaps at most 0.5 percentage points. Its 0.612-point
utility gain supports a local benefit over FULM-MEAN;
the other pairs have larger residual gaps. For fixed centroids,
MEAN at $\gamma$ equals SUM at $\gamma/3$ in real arithmetic,
so averaging itself changes scale rather than update direction.

\subsection{Support- and Norm-Matched Weighting Control}
\label{app:weight_control}
A separate 20-evaluation control compares continuous Fisher
weights with uniform weights on identical Fisher-selected
Top-10\% masks, preserving clients, task vectors, and
forget/retain aggregation and denominator normalization.
Neither method uses QP. The non-Fisher update is globally
rescaled to match the Fisher update's $\ell_2$ norm; both
use $\gamma=0.5,0.6,\ldots,1.4$. Selection-mask mismatches
are zero, and the maximum relative norm mismatch after
BF16 rounding is 0.00212\%.
At $\gamma=1.1$, Bio accuracy is identical (42.97\%), while
Fisher/uniform Cyber accuracy is 32.76/32.56\% and MMLU
is 49.76/49.94\%. Fisher has localized utility gains at
nearest Cyber matches, but these do not also match Bio;
neither method dominates the observed frontiers.
This control does not establish a consistent advantage
for continuous Fisher weighting alone. It is distinct
from dense averaging, which also changes sparse support.

\subsection{Retention-Tolerance Sensitivity}
Table~\ref{tab:sweepe} preserves the separately recorded
fixed-$\gamma$ tolerance sweep. Its no-QP Bio/Cyber/MMLU
reference is 42.97/31.96/48.16\%, whereas the original main
sweep in Table~\ref{tab:sweep} records 42.42/31.91/48.29\%
at the same $\gamma=1$. These are distinct recorded evaluations;
the available records do not establish the exact source of
the discrepancy. Projection effects are compared within the
tolerance sweep, not against the main-table reference.
Tighter budgets improve utility while weakening forgetting;
the regenerated matched-forgetting comparisons below assess
benefits beyond this trade-off.

\begin{table}[!htbp]
\centering
\small
\caption{Retention tolerance~$\eps$ effects on WMDP
(Zephyr-7B, $\gamma=1.0$). Smaller $\eps$ enforces a tighter per-tensor proxy budget.}
\label{tab:sweepe}
\begin{tabular}{c|ccc}
\toprule
$\eps$ & Bio$\downarrow$ & Cyber$\downarrow$ & Util$\uparrow$ \\
\midrule
$\infty$          & \textbf{42.97} & \textbf{31.96} & 48.16 \\
$1{\times}10^{-4}$   & 49.57 & 36.39 & 51.68 \\
$6{\times}10^{-5}$   & 50.75 & 37.49 & 52.41 \\
$4.1{\times}10^{-5}$ & 52.00 & 37.85 & 52.85 \\
$2{\times}10^{-5}$   & 54.05 & 38.35 & 53.29 \\
$1{\times}10^{-5}$   & 55.07 & 38.85 & \textbf{53.45} \\
\bottomrule
\end{tabular}
\end{table}

\subsection{Matched Projection Comparisons}
Table~\ref{tab:projection_selected} combines QP/base and
tighter/standard-QP comparisons on regenerated checkpoints.
It includes every nontrivial jointly close match from the
recorded analyses: both accuracy gaps must be at most 0.5
percentage points, and the QP/base update must be active.
The two inactive QP/base pairs ($\gamma=0.1,0.2$) are omitted;
each has zero accuracy and utility differences.

QP/base references minimize the maximum Bio/Cyber accuracy
gap over the 17 base settings. For the 22 tighter-QP settings,
nearest standard-QP references are selected separately by
domain, breaking ties by nearest $\gamma$, then lower $\gamma$;
only pairs selecting the same reference in both domains are
included. Neither rule uses MMLU. Negative utility differences
are retained. These local comparisons do not establish uniform
dominance. Finite $\eps$ is applied separately to each tensor.

\makeatletter
\setlength{\@fptop}{0pt}
\makeatother
\begin{table}[!tp]
\centering
\small
\setlength{\tabcolsep}{3pt}
\caption{Jointly close projection comparisons on WMDP. F: base; Q: $\eps=4.1\times10^{-5}$; T: $\eps=10^{-5}$. Parentheses give first/second $\gamma$. Gaps are absolute; $\Delta$MMLU is first minus second, in percentage points. Selection uses forgetting gaps, not utility.}
\label{tab:projection_selected}
\begin{tabular}{lrrr}
\toprule
Pair ($\gamma$) & Bio gap & Cyber gap & $\Delta$MMLU \\
\midrule
Q/F (0.3/0.3) & 0.157 & 0.050 & -0.028 \\
Q/F (0.4/0.4) & 0.393 & 0.252 & -0.036 \\
Q/F (0.5/0.5) & 0.393 & 0.101 & -0.114 \\
Q/F (0.7/0.6) & 0.157 & 0.151 & -0.085 \\
Q/F (1.5/1.1) & 0.000 & 0.201 & +0.669 \\
Q/F (2.2/1.5) & 0.393 & 0.302 & +1.723 \\
Q/F (2.4/1.6) & 0.314 & 0.201 & +0.242 \\
\midrule
T/Q (1.1/1.0) & 0.079 & 0.101 & +0.249 \\
T/Q (1.5/1.4) & 0.471 & 0.252 & +0.349 \\
T/Q (2.4/2.3) & 0.393 & 0.302 & +1.453 \\
T/Q (2.6/2.4) & 0.079 & 0.000 & -0.883 \\
\bottomrule
\end{tabular}
\end{table}

\end{document}